\documentclass{article} % For LaTeX2e
\usepackage{nips13submit_e,times}
\usepackage{hyperref}
\usepackage{url}
\usepackage{natbib}
\usepackage{amsmath}
\usepackage{amssymb}
\usepackage{graphicx}

\title{An empirical analysis of dropout \\ in piecewise linear networks}

\author{
David Warde-Farley,
\ \  Ian J. Goodfellow,\ \  Aaron Courville,\ \  Yoshua Bengio\\
D\'epartement d'informatique et de recherche op\'erationnelle\\
Universit\'e de Montr\'eal\\
Montr\'eal, QC H3C 3J7 \\
\texttt{\{wardefar,goodfeli\}@iro.umontreal.ca},\\ \texttt{\{aaron.courville,yoshua.bengio\}@umontreal.ca}\\
}

\nipsfinalcopy % Uncomment for camera-ready version

\begin{document}

\maketitle

\begin{abstract}
The recently introduced dropout training criterion for neural networks has been
the subject of much attention due to its simplicity and remarkable
effectiveness as a regularizer, as well as its interpretation as a training
procedure for an exponentially large ensemble of networks that share
parameters. In this work we empirically investigate several questions related
to the efficacy of dropout, specifically as it concerns networks employing the
popular rectified linear activation function. We investigate the quality of the
test time weight-scaling inference procedure by evaluating the geometric
average exactly in small models, as well as compare the performance of the
geometric mean to the arithmetic mean more commonly employed by ensemble
techniques. We explore the effect of tied weights on the ensemble
interpretation by training ensembles of masked networks without tied weights.
Finally, we investigate an alternative criterion based on a biased estimator of
the maximum likelihood ensemble gradient.
\end{abstract}
\section{Introduction}

Dropout~\citep{Hinton-et-al-arxiv2012} has recently garnered much attention as
a novel regularization strategy for neural networks involving the use of
structured masking noise during stochastic gradient-based optimization. Dropout
training can be viewed as a form of ensemble learning similar to bagging
\citep{ML:Breiman:bagging} on an ensemble of size exponential in the number of
hidden units and input features, where all members of the ensemble share
subsets of their parameters. Combining the predictions of this enormous
ensemble would ordinarily be prohibitively expensive, but a scaling of the
weights admits an approximate computation of the geometric mean of the ensemble
predictions.

Dropout has been a crucial ingredient in the winning solution to several
high-profile competitions, notably in visual object recognition
\citep{Krizhevsky-2012} as well as the Merck Molecular Activity Challenge and
the Adzuna Job Salary Prediction competition. It has also inspired work on
activation function design~\citep{Goodfellow+al-ICML2013-small} as well as
extensions to the basic dropout technique
~\citep{Wan+al-ICML2013-small,WangManning-ICML2013-small} and similar fast
approximate model averaging methods ~\citep{Zeiler+et+al-arxiv2013}. 

Several authors have recently investigated the mechanism by which dropout
achieves its regularization effect in linear
models~\citep{Baldi+al-2013,WangManning-ICML2013-small,Wager+al-2013}, as well
as linear and sigmoidal hidden units~\citep{Baldi+al-2013}. However, many of
the recent empirical successes of dropout, and feed forward neural networks
more generally, have utilised piecewise linear activation
functions~\citep{Jarrett-ICCV2009,Glorot+al-AI-2011,Goodfellow+al-ICML2013-small,Zeiler+al-ICASSP-2013}.
In this work, we empirically study dropout in \textit{rectified linear} networks, employing the recently popular 
hidden unit activation function $f(x) = \max(0, x)$.

We begin by expanding upon previous work which investigated the quality of dropout's approximate ensemble prediction by comparing against Monte
Carlo estimates of the correct geometric average~\citep{Srivastava-master-small,Goodfellow+al-ICML2013-small}.
Here, we compare against the true average, in networks of size small enough
that the exact computation is tractable. We find, by exhaustive enumeration of
all sub-networks in these small cases, that the weight scaling approximation
is a remarkably and somewhat surprisingly accurate surrogate for the true geometric mean.

Next, we consider the importance of the geometric mean itself.
Traditionally, bagged ensembles produce an averaged prediction via the
arithmetic mean, but the weight scaling trick employed with dropout provides an
efficient approximation only for the geometric mean. While, as noted by
\citep{Baldi+al-2013}, the difference between the two can be bounded
\citep{Cartwright+Field-1978}, it is not immediately obvious what effect this
source of error will have on classification performance in practice. We
therefore investigate this question empirically and conclude that the geometric
mean is indeed a suitable replacement for the arithmetic mean in the context of
a dropout-trained ensemble.

The questions raised thus far pertain primarily to the approximate model
averaging performed at test time, but {\em dropout training} also raises
some important questions. At each update, the dropout learning rule follows the
same gradient that true bagging training would follow. However, in the case of
traditional bagging, all members of the ensemble would have independent
parameters. In the case of dropout training, all of the models share
subsets of their parameters. It is unclear how much this coordination serves
to regularize the eventual ensemble. It is also not clear whether the most
important effect is that dropout performs model averaging, or that dropout
encourages each individual unit to work well in a variety of contexts.

To investigate this question, we train a set of independent models
on resamplings (with replacement) of the training data, as in traditional
bagging. Each ensemble member is trained with a single randomly sampled dropout
mask fixed throughout all steps of training. We combine these independently
trained networks into ensembles of varying size, and compare the ensembles'
performance with that of a single network of identical size, trained instead
with dropout. We find evidence to support the claim that the weight sharing
taking place in the context of dropout (between members of the implicit
ensemble) plays an important role in further regularizing the ensemble.

Finally, we investigate an alternative criterion for training the exponentially
large shared-parameter ensemble invoked by dropout. Rather than performing
stochastic gradient descent on a randomly selected sub-network in a manner
similar to bagging, we consider a biased estimator of the gradient of the
geometrically averaged ensemble log likelihood (i.e. the gradient of the
model being approximately evaluated at test-time), with the particular
estimator bearing a resemblance to boosting~\citep{ML:Schapire:weaklearn}. We
find that this new criterion, employing masking noise with the exact same
distribution as is employed by dropout, yields no discernible robustness gains
over networks trained with ordinary stochastic gradient descent. 

\section{Review of dropout}

Dropout is an ensemble learning and prediction technique that can be applied to
deterministic feedforward architectures that predict a target $y$ given input
vector $v$. These architectures contain a series of hidden layers
$\mathbf{h}=\{h^{(1)}, \dots, h^{(L)}\}$.  Dropout trains an ensemble of models
consisting of the set of all models that contain a subset of the variables in
both $v$ and $\mathbf{h}$. The same set of parameters $\theta$ is used to
parameterize a family of distributions $p(y \mid v ; \theta, \mu)$ where $\mu
\in \mathcal{M}$ is a binary mask vector determining which variables to include
in the model, e.g., for a given $\mu$, each input unit and each hidden unit is
set to zero if the corresponding element of $\mu$ is 0.  On each presentation
of a training example, we train a different sub-network by following the gradient
of $\log p(y \mid v; \theta, \mu)$ for a different randomly sampled $\mu$. For
many parameterizations of $p$ (such as most multilayer perceptrons) the
instantiation of different sub-networks $p(y \mid v; \theta, \mu)$ can be
obtained by element-wise multiplication of $v$ and $\mathbf{h}$  with the mask
$\mu$.

\subsection{Dropout as bagging}

Dropout training is similar to bagging~\citep{ML:Breiman:bagging} and related ensemble methods~\citep{Opitz+Maclin-1999}. 
Bagging is an ensemble learning technique in which a set of
models are trained on different subsets of the same dataset. At test time, the predictions of each of the models are averaged
together. The ensemble predictions formed by voting in this manner tend to generalize better than the predictions of the individual
models.

Dropout training differs from bagging in three ways:

\begin{enumerate}
\item All of the models share parameters. This means that they are no longer really trained on separate subsets of the dataset, and
much of what we know about bagging may not apply.
\item Training stops when the ensemble starts to overfit. There is no guarantee that the individual models will be trained to convergence.
In fact, typically, the vast majority of sub-networks are never trained for even one gradient step.
\item Because there are too many models to average together explicitly, dropout averages them together with a fast approximation.
This approximation is to the geometric mean, rather than the arithmetic mean.
\end{enumerate}

\subsection{Approximate model averaging}

The functional form of the model becomes important when it comes time for the ensemble to make a prediction by averaging together all the sub-networks' predictions.
When $p(y \mid v ; \theta) = \mathrm{softmax}(v^T W + b)$, the predictive distribution defined by renormalizing the geometric mean of $p(y \mid v ; \theta, \mu)$ over $\mathcal{M}$ is simply given by $\mathrm{softmax}(v^T W /2 + b)$.
This is also true for sigmoid output units, which are special cases of the
softmax. This result holds exactly in the case of a single layer softmax model~\citep{Hinton-et-al-arxiv2012} or an MLP with no non-linearity applied to each unit~\citep{Goodfellow+al-ICML2013-small}.
Previous work on dropout applies the same scheme in deep architectures with
hidden units that have nonlinearities, such as rectified linear units, where
the $W/2$ method is only an approximation to the geometric mean.
The approximation has been characterized mathematically for linear and sigmoid networks~\citep{Baldi+al-2013,Wager+al-2013},
but seems to perform especially well in practice for nonlinear networks with piecewise linear activation functions~\citep{Srivastava-master-small,Goodfellow+al-ICML2013-small}.

\section{Experimental setup}

Our initial investigations employed rectifier networks with 2 hidden layers and
10 hidden units per layer, and a single logistic sigmoid output unit. We
applied this class of networks to six binary classification problems
derived from popular multi-class benchmarks, simplified in this fashion in
order to allow for much simpler architectures to effectively solve the task,
as well as a synthetic task of our own design.

Specifically, we chose four binary sub-tasks from the MNIST handwritten digit
database~\citep{LeCun+98}. Our training sets consisted of all occurrences of
two digit classes (1 vs. 7, 1 vs. 8, 0 vs. 8, and 2 vs. 3) within the first
50,000 examples of the MNIST training set, with the occurrences from the last
10,000 examples held back as a validation set. We used the corresponding
occurrences from the official MNIST test set for evaluating test error. 

We also chose two binary sub-tasks from the CoverType dataset of the UCI
Machine Learning Repository, specifically discriminating classes 1 and 2
(Spruce-Fir vs. Lodgepole Pine) and classes 3 and 4 (Ponderosa Pine vs.
Cottonwood/Willow).  This task represents a very different domain than the
first two datasets, but one where neural network approaches have nonetheless
seen success (see e.g. ~\citet{Dauphin-et-al-NIPS2011}).\footnote{Unlike
\citet{Dauphin-et-al-NIPS2011}, we train and evaluate on the records of each
class from the data split advertised in the original dataset description.
This makes the task much more challenging and many methods prone to overfitting.}

The final task is a synthetic task in two dimensions: inputs lie in $(-1, 1)
\times (-1, 1) \subset \mathbb{R}^2$, and the domain is divided into two
regions of equal area: the diamond with corners $(1, 0)$, $(0, 1)$, $(-1, 0)$,
$(0, -1)$ and the union of the outlying triangles. 
%A randomly generated
%training set is shown in Figure~\ref{diamondtrain}.
In order to keep the
synthetic task moderately challenging, the training set size was restricted to
100 points sampled uniformly at random. An additional 500 points were sampled
for a validation set and another 1000 as a test set.

% \begin{figure}[h]
% \label{diamondtrain}
% \centering
% \includegraphics[width=2.5in]{diamond_train}
% \caption{Training data consisting of 100 points sampled uniformly at random on $[-1, 1]^2$, with label
% corresponding to whether or not the point falls within a diamond with corners (1, 0), (0, 1), (-1, 0) and (0, -1).}
% \end{figure}

%\begin{figure}[h]
%\label{diamondnets}
%\centering
%\includegraphics[width=1.7in]{withoutdrop}
%\includegraphics[width=1.7in]{withdrop}
%\caption{(Left) The decision boundary learned by a 2-10-10-1 rectifier
%network with a sigmoid output, without dropout, on the training set in Figure~\ref{diamondtrain}. (Right) the decision boundary learned by the same architecture with dropout. Both networks were trained with early stopping on the same validation set of another 500 points.}
%\end{figure}

In order to keep the mask enumeration tractable in the case of the larger input
dimension tasks, we chose to apply dropout in the hidden layers only. This has
the added benefit of simplifying the ensemble computation: though dropout is
typically applied in the input layer, inclusion probabilities higher than 0.5
are employed (e.g. $0.8$ in
\citet{Hinton-et-al-arxiv2012,Krizhevsky-2012-small}), making it necessary to
unevenly weight the terms in the average. We chose hyperparameters by random
search~\citep{Bergstra+Bengio-2012-small} over learning rate and momentum
(initial values and decrease/increase schedules, respectively), as well as mini-batch
size.  We performed early stopping on the validation set, terminating when a
lower validation error had not been observed for 100 epochs; when training with
dropout, the figure of merit for early stopping was the
validation error using the weight-scaled predictions.

\section{Weight scaling versus Monte Carlo or exact model averaging}
\label{sec:model_averaging}
\citet{Srivastava-master-small,Goodfellow+al-ICML2013-small} previously
investigated the fidelity of the weight scaling approximation in the context of
rectifier networks and maxout networks, respectively, through the use of a
Monte Carlo approximation to the true model average. By concerning ourselves
with small networks where exhaustive enumeration is possible, we were able to
avoid the effect of additional variance due to the Monte-Carlo average and
compute the exact geometric mean over all possible dropout sub-networks.

On each of the 7 tasks, we randomly sampled 50 sets of hyperparameters and
trained 50 networks with dropout. We then computed, for each point in the
test set for each task, the activities of the network corresponding to
each of the $2^{20}$ possible dropout masks. We then geometrically averaged
their predictions (by arithmetically averaging all values of the input to
the sigmoid output unit) and computed the geometric average prediction for
each point in the test set. Finally, we compared the misclassification rate
using these predictions to that obtained using the approximate, weight-scaled
predictions.
\begin{figure}[ht]
\centering
\includegraphics[width=4.2in]{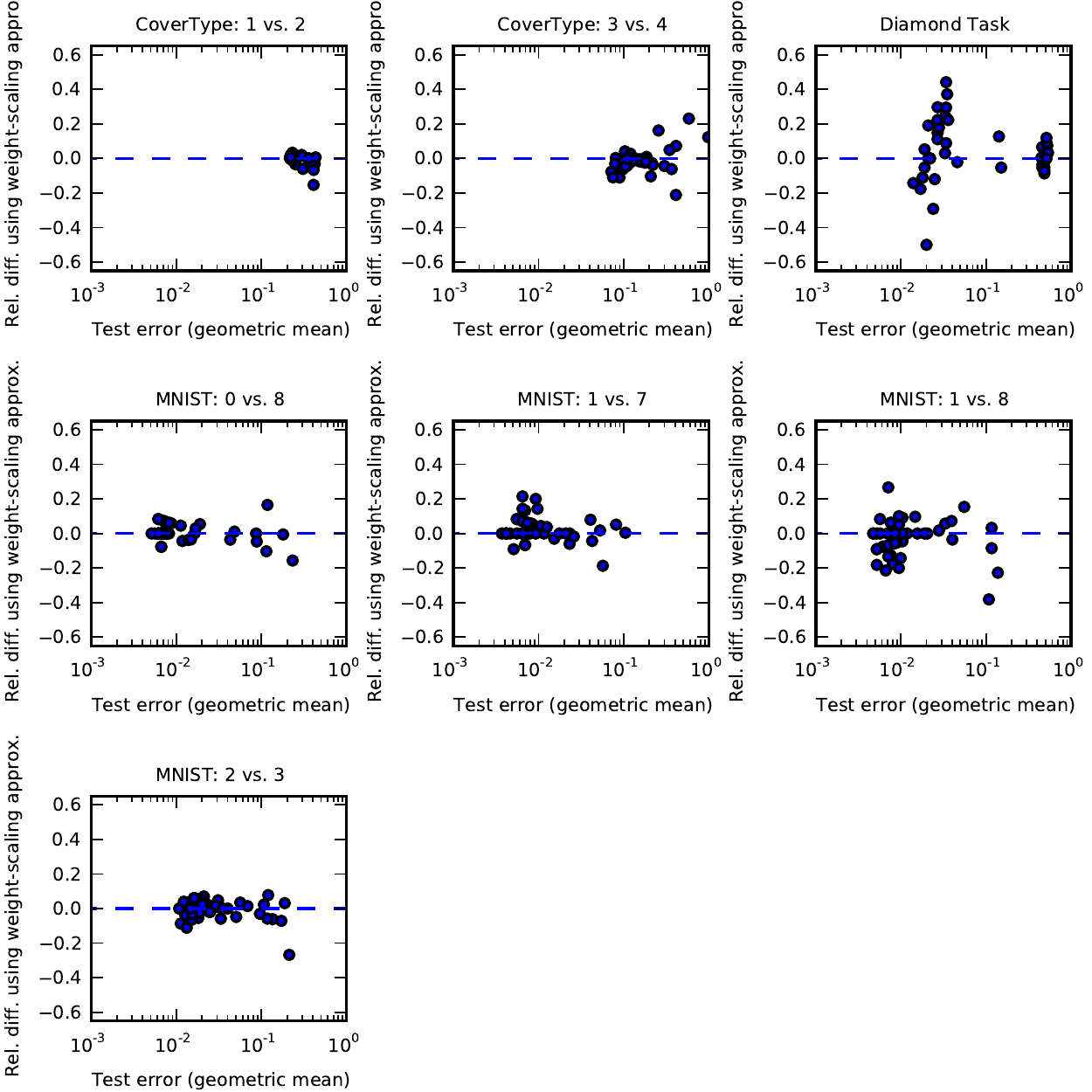}
\caption{\small Comparison of test error obtained with an exhaustive computation of the geometric mean (on the $x$-axes) and the relative difference in the test error obtained with the weight-scaling approximation.}
\label{fig:scaling}
\end{figure}

The results are shown in Figure~\ref{fig:scaling}, where each point
represents a different hyperparameter configuration. The overall result is that
the approximation yields a network that performs very similarly.  In order to
make differences visible, we plot on the $y$-axis the relative difference in
test error between the true geometric average network and the weight-scaled
approximation for different networks achieving different values of the test
error.

Additionally, we statistically tested the fidelity of the approximation via the
Wilcoxon signed-rank test, a nonparametric paired sample test similar to the
paired $t$-test, applying a Bonferroni correction for multiple hypotheses.  At
$\alpha = 0.01$, no significant differences were observed for any of the seven
tasks.

\section{Geometric mean versus arithmetic mean}

Though the inexpensive computation of an approximate geometric mean was noted
in~\citep{Hinton-et-al-arxiv2012}, little has been said of the choice of the
geometric mean. Ensemble methods in the literature often employ an arithmetic
mean for model averaging. It is thus natural to pose the question as to whether
the choice of the geometric mean has an impact on the generalization
capabilities of the ensemble.

Using the same networks trained in Section~\ref{sec:model_averaging}, we
combined the forward-propagated predictions of all $2^{20}$ models using the
arithmetic mean. \begin{figure}[t]
\centering
\includegraphics[width=4.2in]{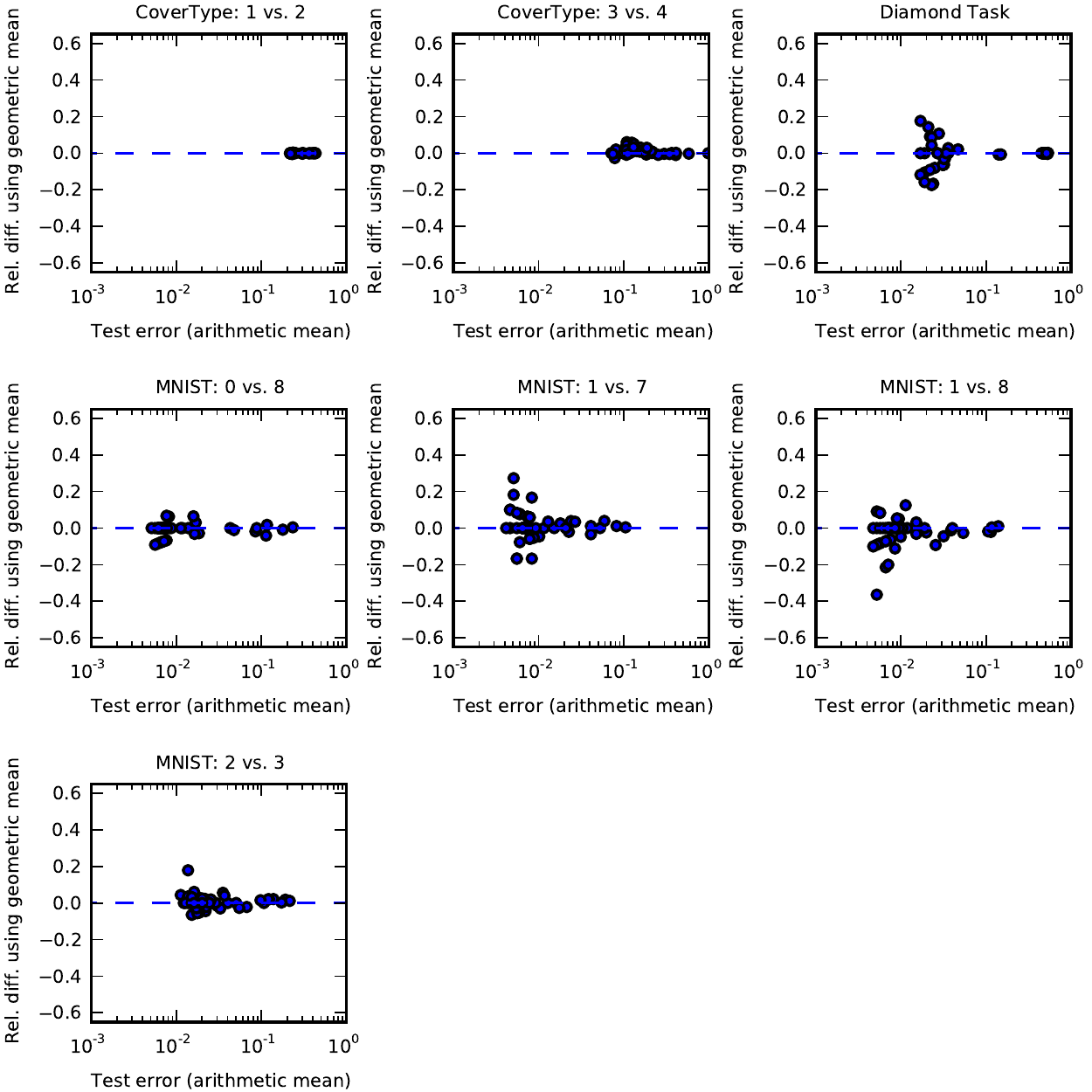}
\caption{\small Comparison of test error obtained with an exhaustive computation of the arithmetic mean (on the $x$-axes) and the relative difference in the test error obtained with the (exhaustively computed) geometric mean.}
\label{fig:ari_vs_geo}
\end{figure}
In Figure~\ref{fig:ari_vs_geo}, we plot the relative
difference in test error between the arithmetic mean predictions. We find that
across all seven tasks, the geometric mean is a reasonable proxy for the
arithmetic mean, with relative error rarely exceeding 20\% except for the
synthetic task. In absolute terms, the discrepancy between the
test error achieved by the geometric mean and the arithmetic mean never
exceeded 0.75\% for any of the tasks.

\section{Dropout ensembles versus untied weights}
\label{sec:untied}

We now turn from our investigation of the characteristics of inference in
dropout-trained networks to an investigation of the training procedure.
For the remainder of the experiments, we trained networks of a more realistic
size and capacity on the full multiclass MNIST problem. Once again, we
employed two layers of rectified linear units. In addition to dropout, we
utilised norm constraint regularization on the incoming weights to each
hidden unit. We again performed random search over hyperparameter values, now
including in our search the initial ranges of weights, the number of hidden
units in each of two layers, and the maximum weight vector norms
of each layer.

Dropout training can be viewed as performing bagging on an ensemble that is of
size exponential in the number of hidden units, where each member of the
ensemble shares parameters with other members of the ensemble. Because each
gradient step is taken on a different mini-batch of training data, each
sub-network can be seen to be trained on a different resampling of the training
set, as in traditional bagging. Furthermore, while each step is taken with
respect to the log likelihood of a single ensemble member, the effect of the
weight update is applied to all members of the ensemble
simultaneously\footnote{At least, all members of the ensemble that share any
parameters with the sub-network just updated. There certainly exist pairs of
ensemble members whose parameter sets are disjoint.} We investigate the role of
this complex weight-sharing scheme by training an ensemble of independent
networks on resamplings of the training data, each with a single dropout mask
fixed in place throughout training.

We first performed a hyperparameter search by sampling 50 hyperparameter
configurations and choosing the network with the lowest validation error. The
best of these networks obtains a test error of 1.06\%, matching results
reported by~\citet{Srivastava-master-small}. Using the same hyperparameters,
we trained 360 models initialized with different random seeds, on different
resamplings (with replacement) of the training set, as in traditional
bagging. Instead of applying dropout during training (and thus applying a
different mask at each gradient step), we sampled one dropout mask per model
and held it fixed throughout training and at test time. The resulting networks
thus have architectures sampled from the same distribution as the sub-networks
trained during dropout training, but each network's parameters are independent
of all other networks.

\begin{figure}[h]
\centering
\includegraphics[width=2.7in]{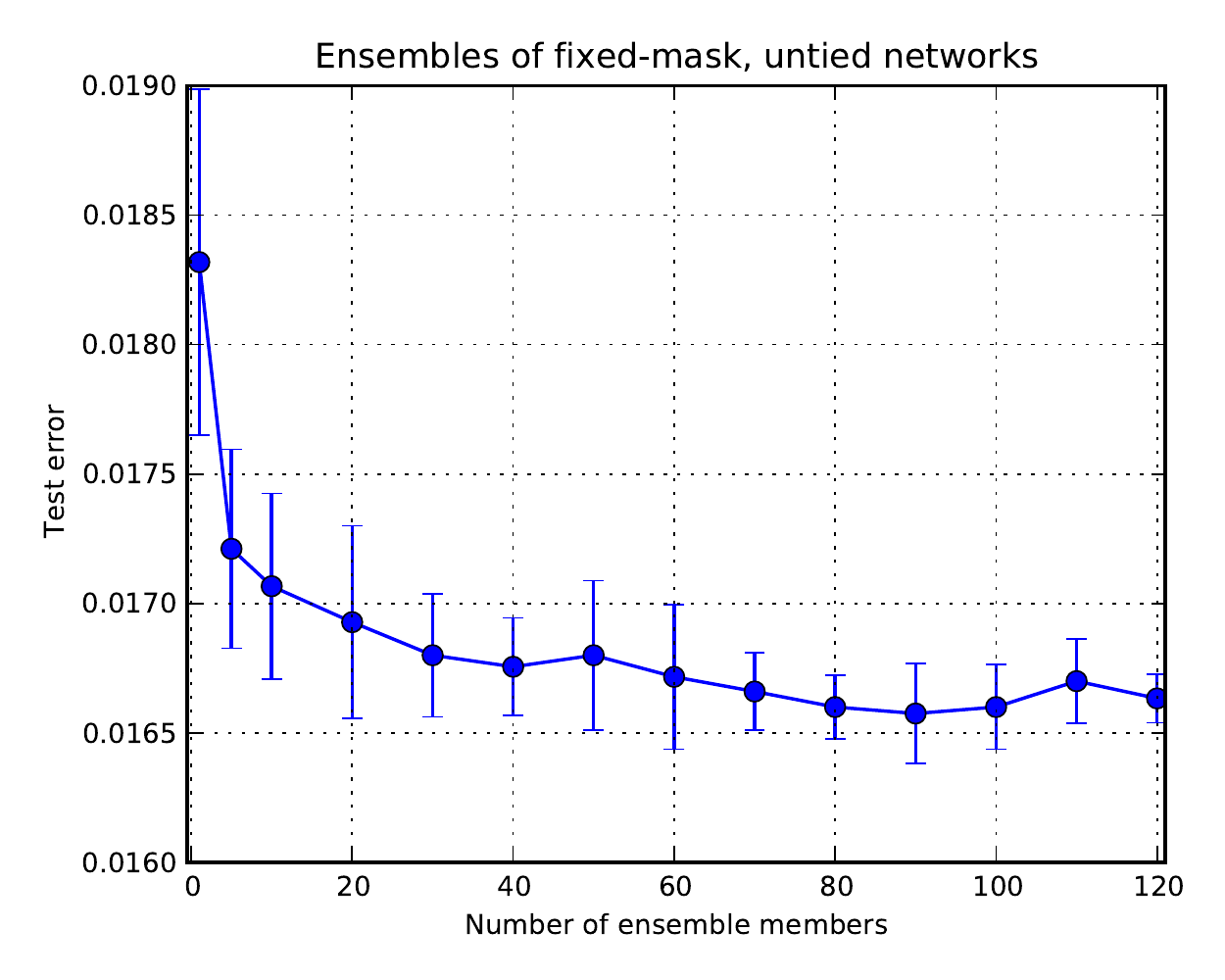}
\caption{\small Average test error on MNIST for varying sizes of untied-weight
ensembles. 360 networks were trained to convergence, each with a single
randomly sampled dropout mask fixed in place throughout. These networks
pre-softmax activations were then averaged to produce predictions for varying
sizes of ensembles. For each size $n$, $\lfloor360/n\rfloor$ disjoint subsets
were combined in this fashion, and the test error mean and standard deviation
over ensembles is shown here.}
\label{fig:untied_ensemble}
\end{figure}

We then evaluate test error for ensembles of these networks, combining their
predictions (with the dropout mask used during training still fixed in place
at test time) via the geometric mean, as is approximately done in the context
of dropout. Our results for various sizes of ensemble are shown in
Figure~\ref{fig:untied_ensemble}.
Our results suggest that there indeed an effect; combining all 360
independently trained models yields a test error of 1.66\%, far above the
even the suboptimally tuned networks trained with dropout.
Aside from the size of the independent ensemble being considerably smaller,
one potential confounding factor is that the non-architectural hyperparameters
were selected in the context of their performance when using dropout and used
as-is to train the networks with untied weights; although each of these was
early-stopped independently, it remains unclear how to efficiently
optimize hyperparameters for the individual members of a large ensemble so
as to facilitate a fairer comparison (indeed, this highlights a general issue
with the high cost of training ensembles of neural networks, that dropout
conveniently sidesteps).

\section{Dropout bagging versus dropout boosting}

Other algorithms such as denoising autoencoders~\citep{Vincent-JMLR-2010} are
motivated by the idea that models trained with noise are robust to slight
transformations of their inputs. Previous work has drawn connections
between noise and regularization penalties~\citep{bishop95training}; similar
connections in the case of dropout have recently been noted
\citep{Baldi+al-2013,Wager+al-2013}. It is natural to question whether dropout
can be wholly characterized in terms of learned noise robustness, and whether
the model-averaging perspective is necessary or fruitful.

In order to investigate this question we propose an algorithm that injects
exactly the same noise as dropout. For this test to be effective, we require an
algorithm that can successfully minimize training error, and obtain acceptable
generalization performance. It needs to perform at least as well as standard
maximum likelihood; otherwise all we have done is designed a pathological
algorithm that fails to train.

We therefore introduce {\em dropout boosting}. The objective function for each
(sub-network, example) pair in dropout boosting is the likelihood of the data
according to the ensemble; however, only the parameters of the current
sub-network may be updated for each example. Ordinary dropout performs bagging
by maximizing the likelihood of the correct target for the current example {\em
under the current sub-network}, whereas dropout boosting takes into account the
contributions of other sub-networks, in a manner reminiscent of boosting.

The objective function for dropout is
$\frac{1}{2^{|\mathcal{M}|}} \sum_{\mu \in \mathcal{M}} \log p( y \mid v; \theta, \mu)$.
For dropout boosting, assume each mask $\mu$ has a separate set of parameters $\theta_\mu$
(though in reality these parameters are tied, as in conventional dropout). The dropout boosting objective function
is then given by $\log p_{\text{ensemble}}( y \mid v; \theta)$,
where 
\[ p_{\text{ensemble}} ( y \mid v ; \theta) = \frac{1}{Z} \tilde{p}(y \mid v ; \theta) \]
\[ Z = \sum_{y'} \tilde{p} ( y' \mid v ; \theta) \]
\[ \tilde{p} ( y \mid v ; \theta) = \sqrt[2^{|\mathcal{M}|}]{\Pi_{\mu \in \mathcal{M}} p(y \mid v; \theta_\mu) }. \]
The boosting learning rule is to select one model and update its parameters
given all of the other models. In conventional boosting, these other models have
already been trained to convergence. In dropout boosting, the other models
actually share parameters with the network being trained at any given step, and
initially the other models have not been trained at all. The learning rule is
to select a sub-network indexed by $\mu$ and follow the ensemble gradient 
$\nabla_{\theta_\mu} \log p_{\text{ensemble}} ( y \mid v ; \theta )$, i.e.
\[ \Delta \theta_\mu \propto \frac {1} {2 ^ {|\mathcal{M}|} } \left(
\nabla_{\theta_\mu} \log p( y \mid v ; \theta_\mu, \mu)
+ \sum_{y'} p_{\text{ensemble}}(y' \mid v ) \nabla_{\theta_\mu} \log p( y' \mid v; \theta_\mu, \mu) 
\right).\]

% Thus far this appears simply to be gradient descent on the ensemble likelihood,^
% which is intractable due to the exponential sum that appears in the gradient
% of $Z$. Several Monte Carlo estimators of the gradient can be formulated; the
% estimator that we discovered to work best in practice, which we term ``dropout
% boosting'' due to the particular form it takes, estimates the intractable sum
% with a single term, corresponding to the same mask employed in the 

Rather than using the boosting-like algorithm, one could obtain a generic
Monte-Carlo procedure for maximizing the log likelihood of the ensemble by
averaging together the gradient for multiple values of $\mu$, and optionally
using a different $\mu$ for the term in the left and the term on the right.
Empirically, we obtained the best results in the special case of boosting,
where the term on the left uses the same $\mu$ as the term on the right -- that
is, both terms of the gradient apply updates only to one member of the
ensemble, even though the criterion being optimized is global.

Note that the intractable $p_{\text{ensemble}}$ still appears in the learning
rule.  To implement the training algorithm efficiently, we can approximate the
ensemble predictions using the weight scaling approximation. This introduces
further bias into the estimator, but our findings in Section
\ref{sec:model_averaging} suggest that the approximation error is small.

Note that dropout boosting employs exactly the same noise as regular dropout
uses to perform bagging, and thus should perform similarly to conventional
dropout if learned noise robustness is the important ingredient. If we instead take
the view that this is a large ensemble of complex learners whose likelihood
is being jointly optimized, we would expect that employing a criterion more
similar to boosting than bagging would perform more poorly.
As boosting maximizes the likelihood of the ensemble, it would perhaps
be prone to overfitting in this setting, as the ensemble is very large and
the learners are not particularly weak.

\begin{figure}[h]
\centering
\includegraphics[width=3.9in]{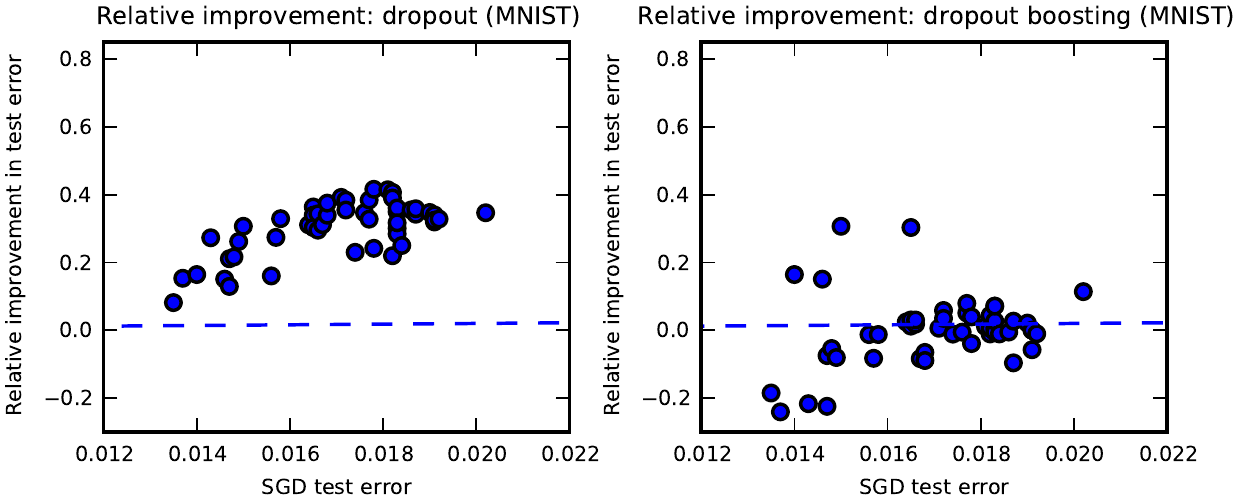}
\caption{\small Comparison of dropout (left) and dropout boosting (right) to stochastic gradient descent with matched hyperparameters.}
\label{fig:dropout_boosting}
\end{figure}
Starting with the 50 models trained in Section~\ref{sec:untied}, we employed
the same hyperparameters to train a matched set of 50 networks with dropout
boosting, and another with plain stochastic gradient descent. In Figure~\ref{fig:dropout_boosting}, we plot the relative performance of dropout and dropout boosting compared to a model with the same hyperparameters trained with SGD. While
dropout unsurprisingly shows a very consistent edge, dropout boosting performs,
on average, little better than stochastic gradient descent. The Wilcoxon
signed-rank test similarly failed to find a significant difference between
dropout boosting and SGD ($p > 0.7$). While several outliers approach very good
performance (perhaps owing to the added stochasticity), dropout boosting is, on
average, no better and often slightly worse than maximum likelihood training,
in stark contrast with dropout's systematic advantage in generalization
performance.

\section{Conclusion}

We investigated several questions related to the efficacy of dropout,
focusing on the specific case of the popular rectified linear nonlinearity
for hidden units. We showed that the weight-scaling approximation is a
remarkably accurate proxy for the usually intractable geometric mean over
all possible sub-networks, and that the geometric mean (and thus its
weight-scaled surrogate) compares favourably to the traditionally popular
arithmetic mean in terms of classification performance. We demonstrated that
weight-sharing between members of the implicit dropout ensemble appears to
have a significant regularization effect, by comparing to analogously trained
ensembles of the same form that did not share parameters. Finally, we
demonstrated that simply adding noise, even noise with identical
characteristics to the noise applied during dropout training, is not sufficient
to obtain the benefits of dropout, by introducing dropout boosting, a
training procedure utilising the same masking noise as conventional dropout,
which successfully trains networks but loses dropout's benefits, instead
performing roughly as well as ordinary stochastic gradient descent.

Our results suggest that dropout is an extremely effective ensemble learning
method, paired with a clever approximate inference scheme that is remarkably
accurate in the case of rectified linear networks. Further research
is necessary to shed more light on the model averaging interpretation of
dropout. \citet{Hinton-et-al-arxiv2012} noted that dropout forces each hidden
unit to perform computation that is useful in a wide variety of contexts.
Our results with a sizeable ensemble of independent bagged models seem to lend
support to this view, though our experiments were limited to ensembles of
several hundred networks at most, tiny in comparison with the weight-sharing
ensemble invoked by dropout. The relative importance of the astronomically
large ensemble versus the learned ``mixability'' of hidden units remains an
open question.  Another interesting direction involves methods that are able to
efficiently, approximately average over different classes of model that share
parameters in some manner, rather than merely averaging over members of the
same model class.

\subsubsection*{Acknowledgments}

The authors would like to acknowledge the efforts of the many developers of
Theano~\citep{bergstra+al:2010-scipy,Bastien-Theano-2012},
pylearn2~\citep{pylearn2_arxiv_2013} which were utilised in experiments. We
would also like to thank NSERC, Compute Canada, and Calcul Qu\'ebec for
providing computational resources. Ian Goodfellow is supported by the 2013
Google Fellowship in Deep Learning.

\small{
\bibliography{strings,strings-shorter,ml,aigaion}
\bibliographystyle{natbib}

\end{document}